%% file: 00_main.tex
\documentclass[letterpaper, 10 pt, conference]{ieeeconf}  % Comment this line out if you need a4paper

\IEEEoverridecommandlockouts                              % This command is only needed if 
\usepackage{graphicx}
\usepackage{svg}
\usepackage{amsmath}
\usepackage{amssymb}

\title{\LARGE \bf
Tendon Force Modeling for Sim2Real Transfer of Reinforcement Learning Policies for Tendon-Driven Robots
}

\author{Valentin Yuryev$^{1*}$ and Josie Hughes$^{1}$% <-this % stops a space
\thanks{$^{1}$These authors are with with CREATE Lab, Swiss Federal Institute of Technology Lausanne (EPFL), Lausanne,
Switzerland.}%
\thanks{*Correspondence to: V.Yuryev ({\tt\small valentin.yuryev@epfl.ch})}% <-this % stops a space
}

\begin{document}

\maketitle
\thispagestyle{empty}
\pagestyle{empty}

%%%%%%%%%%%%%%%%%%%%%%%%%%%%%%%%%%%%%%%%%%%%%%%%%%%%%%%%%%%%%%%%%%%%%%%%%%%%%%%%
\begin{abstract}

Robots which make use of soft or compliant interactions often leverage tendon-driven actuation which enables actuators to be placed more flexibly, and compliance to be maintained.
However, controlling complex tendon systems is challenging.
Simulation paired with reinforcement learning (RL) could be enable more complex behaviors to be generated.
Such methods rely on torque and force-based simulation roll-outs which are limited by the sim-to-real gap, stemming from the actuator and system dynamics, resulting in poor transfer of RL policies onto real robots.
To address this, we propose a method to model the tendon forces produced by typical servo motors, focusing specifically on the transfer of RL policies for a tendon driven finger.
Our approach extends existing data-driven techniques by leveraging contextual history and a novel data collection test-bench.
This test-bench allows us to capture tendon forces undergo contact-rich interactions typical of real-world manipulation.
We then utilize our force estimation model in a GPU-accelerated tendon force-driven rigid body simulation to train RL-based controllers.
Our transformer-based model is capable of predicting tendon forces within 3\% of the maximum motor force and is robot-agnostic.
By integrating our learned model into simulation, we reduce the sim-to-real gap for test trajectories by 41\%.
RL-based controller trained with our model achieves a 50\% improvement in fingertip pose tracking tasks on real tendon-driven robotic fingers.
This approach is generalizable to different actuators and robot systems, and can enable RL policies to be used widely across tendon systems, advancing capabilities of dexterous manipulators and soft robots.
\end{abstract}

%%%%%%%%%%%%%%%%%%%%%%%%%%%%%%%%%%%%%%%%%%%%%%%%%%%%%%%%%%%%%%%%%%%%%%%%%%%%%%%%
\section{Introduction}
\input{01_introduction}

\section{Methods}
\input{02_methods}

\section{Experimental Setup}
\input{03_experimental_setup}

\section{Experimental Results}
\input{04_results}

\section{Discussion \& Conclusion}
\input{05_discussion}

% \section{Acknowledgments}
% \input{06_acknowledgments}

%%%%%%%%%%%%%%%%%%%%%%%%%%%%%%%%%%%%%%%%%%%%%%%%%%%%%%%%%%%%%%%%%%%%%%%%%%%%%%%%
\bibliographystyle{IEEEtran}
\bibliography{references,ref_extra}
% \bibliography{references}

\end{document}

%% file: 01_introduction.tex
% Goal - Dexterous manipulation/tendon driven systems, need to be able to leverage Sim2real, need force information. 

% Problem (Existing SoA + Approach): 
% In Sim, generate force control.  Motors typically work in position/dynamical systems which limits transfer - we need to be able to predict tendon force.
% Existing work has addressed this for legged locomotion...bu

% Hypothesis
% (1) Can have better transfer of RL policies when using a simulator that has tendon models, and we can actuation models which can predict tendon force.
% (2) Minor - Using transfomers (i.e. having history/context) in comparison to MLP enbales frictional effects to be better captured, also helping sim2real transfer. 

% Contributions/Demonstrations
% (1) Test bench for tendon driven systems, for capturing inline force data, easy to generalize to different type of tendon driven systems including various fingers/other.
% (2) Show that transform based context models can capture non-linear behaviour actuations better than MLPs
% (3) Show the improvement in transfer of RL transfer, for force controllers.

% This unblocks the many robots using tendons + motors (without force control), for leveraging RL for training policies. 

\label{sec:introduction}

% \begin{figure*}[tbp]
%     \centering
%     \includegraphics[width=0.95\textwidth]{figs/fig_rl_ee_tracking.pdf}
%     % \vspace{-4mm}
%     \caption{Plot of end effector tracking controller trained on pure torque model and transformer-based model. Region A shows how the pure torque model fails to track finger opening due to lack of knowledge to stick friction of the motor, causing it to overshoot and fully open the finger.}
%     \label{fig:rl_tracking}
%     % \vspace{-4mm}
% \end{figure*}

% \begin{figure*}[tbp]

% \begin{figure}[tb]
%     \centering
%     \includegraphics[width=\linewidth]{figs/overview_alt.pdf}
%     \vspace{-4mm}
%     \caption{OVERVIEW 2: RL-based tendon force control for dexterous tendon-driven robots using ML-based tendon force estimation.}
%     \label{fig:block_diagram}
%     \vspace{-4mm}
% \end{figure}

% Problem: Dexterous manipulation/tendon driven systems, need to be able to leverage Sim2real, need force information.

Tendon driven actuation is observed widely in biology, particularly in musculoskeletal animals \cite{BiewenerTendonsLigamentsStructure2008}.
Inspired by this, soft robot arms or compliant, dexterous robot hands can benefit from tendon-driven actuation \cite{GildayEmbodiedManipulationFuture2025a, GildayVisionBasedCollocatedActuationSensing2020}.
Tendon enable high force transmission whilst allowing actuators to be located away from the point of actuation, reducing moving inertia and enabling the incorporation of compliance either inline or directly through the tendon \cite{WangSurveyMainDrive2021}.
One key challenge for tendon-driven systems such as dexterous manipulators is efficiently generating robust controllers that can operate in the real world \cite{SuleimenovComparativeAnalysisModelBased2025}.
Combining simulation and reinforcement learning (RL) for generating control policies has been demonstrated to be a highly effective approach for tasks such as locomotion of quadrupeds. 
The generated controllers can be robust to variations in task and generalizable without relying on the collection of extensive real-world data \cite{OpenAISolvingRubiksCube2019, MikiLearningRobustPerceptive2022a}.
However, these RL pipelines require torque or force information, as the RL-generated policies seek to influence future states and behavior through dynamics (force-driven simulation) rather than purely kinematics, such as position control.
However, most motors at the physical scale and force required for tendon driven dexterous manipulators provide only position control or current-based sensing which has limited accuracy \cite{HandaDeXtremeTransferAgile2024a}.
This limits the use of RL for controller generation as one can not accurately know what forces would the motor apply to provide realistic simulation roll out and sample collection for training.
To leverage RL for tendon-driven dexterous manipulators, we require tendon-driven simulators and accompanying methods which enable position-controlled motors to be used to execute force-aware controllers.

Existing efforts have been made to develop model-based approaches which can estimate the properties of servo motors commonly used in tendon-driven systems, to account for friction-related discrepancies \cite{DuclusaudExtendedFrictionModels2025}.
The approach shows promise capturing the friction-related component, but neglects other causes such as tendon slack, motor latencies and other consumer-grade servo motor causes of modelling inaccuracies.
Other works focused on augmenting the simulation robot model with a transformer-based architecture to improve simulation roll-out accuracy \cite{SerifiTransformerBasedNeuralAugmentation2023a}.
Such methods do well augmenting a robotic simulation to match reality, however are not robot-agnostic, requiring training data for new robotics systems.
One can also attempt to bypass the sim-to-real (sim2real) gap by randomizing the simulation domain, simulating a distribution of systems all varying in mass, kinematics and frictions with the assumption that the real robotic system lies somewhere within that distribution \cite{PengSimtoRealTransferRobotic2018}.
Adjacent fields, such as legged robotics, have also shown the use of torque estimation for simulating robots in RL pipelines \cite{HwangboLearningAgileDynamic2019}.
However, for tendon driven systems with conventional servo motors, torque estimation is not readily available and the dynamics of the system can not be captured by the short history window used in these actuation models.

% However, these methods have not been utilized for tendon-driven robots or for simulation of dexterous manipulators.

\begin{figure}[t!]
    \centering
    \includegraphics[width=\linewidth]{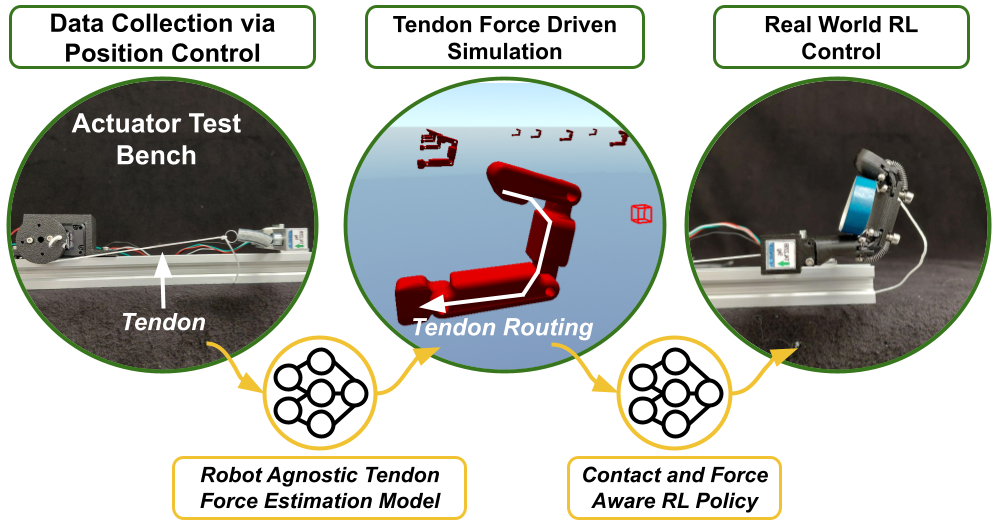}
    % \vspace{-4mm}
    \caption{Overview of our pipeline. We start by collecting tendon force data from real spring mass systems and training an estimator. We then use the estimated forces to simulate accurate tendon forces in highly massively parallelization simulator that supports tendon force driven robots. This allows us to train a contact and tendon force aware reinforcement learning controller that is deployable with minimized sim2real gap.}
    \label{fig:overview}
    % \vspace{-4mm}
% \end{figure*}
\end{figure}

We propose an extension to existing data-driven methods for actuation modeling of tendon-driven systems in order to train RL policies in simulation and transfer to the real world.
To achieve this, we propose tendon-driven setup, shown in Fig. \ref{fig:overview}, for data capture of actuator inline with a mass-spring damper systems, which can recreate full tendon-robot systems.
This enables capturing of general purpose data, but also system specific data, for example for collecting data directly from a tendon-driven dexterous finger both moving freely and under contact.
%To recreate contact-rich manipulation scenarios often seen in dexterous manipulation, we propose a new way to collect data .
To model the actuator-tendon system, we propose a approach which can capture the necessary contextual information from a history buffer, leveraging modern ML architectures such as transformers \cite{VaswaniAttentionAllYou2023, SerifiTransformerBasedNeuralAugmentation2023a}.
Using this actuation model in conjunction with massively parallelizable simulators capable of tendon force driven actuation, we can train RL policies and deploy them on the real system \cite{VsimNextgenPhysics}.
This opens possibilities to training solely motor state-reliant blind proprioceptive controllers for tendon driven systems, leveraging RL for dexterous manipulators, soft robots and more. The main contributions of this work are:
\begin{itemize}
\item A data-driven tendon force estimator for tendon-driven systems that relies only on motor encoder signals, with no force sensors required at inference time, which generalizes across different tendon-driven spring system configurations.
\item Demonstration that force prediction requires temporal context due to actuator dynamics. We show that sequence models such as Transformer encoders achieve more generalizable performance when compared to other architectures.
\item Demonstration that integrating this learned actuator model in simulation reduces the sim-to-real gap for RL policies on tendon-driven systems. We validate this on a real robotic finger with two coupled joints actuated using a single tendon.
\end{itemize}

In the remainder of the paper we first introduce the methods for collecting data, training the actuator model, and using this for training a RL policy in simulation for our tendon driven finger.  We then show experimental results of both the actuator performance and our RL controllers.

%% file: 02_methods.tex
\label{sec:methods}
Our method addresses the problem of sim-to-real transfer for RL policies on tendon-driven robotic systems, focusing on a tendon driven finger. Tendon driven systems have complex and nonlinear actuator dynamics that are difficult to represent analytically, causing control policies trained in simulation requiring torque data, to fail when deployed on real hardware. We propose a solution to this problem by learning a data-driven tendon force model that maps motor encoder signals to tendon forces. This leverages our data-collection test-bench, which incorporates a load-cell inline with the system, which can capture tendon force without modification of the robotic system.
We integrate the learnt tendon force model in simulation to better represent real actuator behavior, and enable real-world translation of RL policies. 

The following sections describe the tendon-driven system dynamics (\ref{sec:system_dynamics}), our data-driven approach to learning the tendon force estimation model (\ref{sec:tendon_model}), the data collection platform (\ref{sec:data_collection}), simulation of our tendon-driven system (\ref{sec:sim}) and the RL policy training domain randomization techniques specific to tendon force driven simulation (\ref{sec:RL}).

\subsection{Tendon-Driven System Problem Definition}
\label{sec:system_dynamics}

\begin{figure}[tb]
    \centering
    \includegraphics[width=\linewidth]{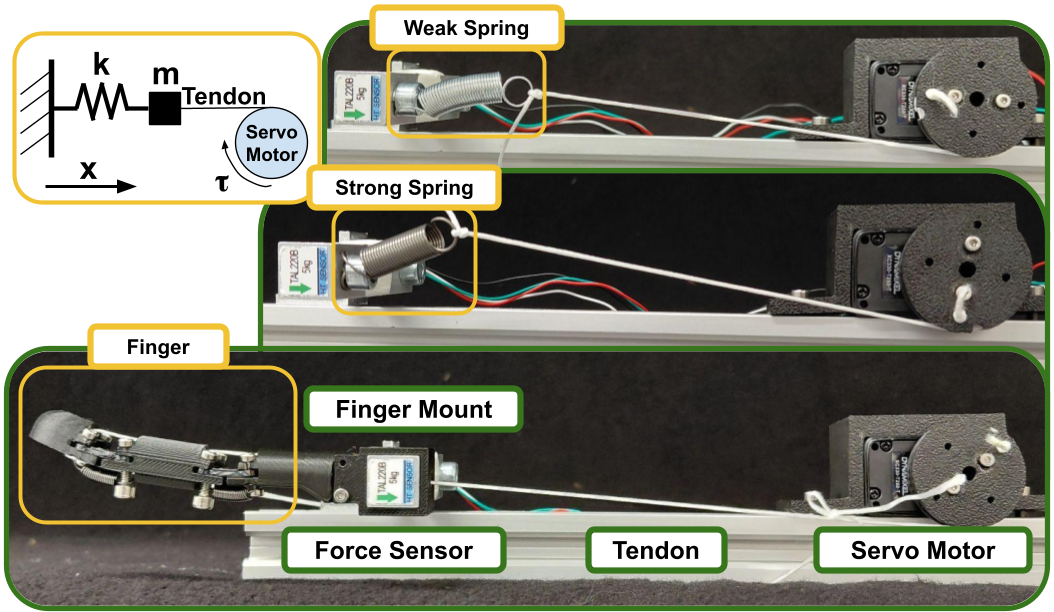}
    \vspace{-4mm}
    \caption{Tendon-driven system setup for data collection using various springs and finger setup. 
    The system can be represented as a mass spring system. 
    The force sensor detects tendon tension forces applied by the servo motor to the system.
    For the finger setup, the tendon is routed directly through the load cell, allowing for data collection directly on a component of a dexterous hand.}
    \label{fig:spring_mass}
    \vspace{-4mm}
\end{figure}

We consider our tendon driven system to consist of a position controlled motor actuating a tendon which transmits force to a mechanical system. 
The system can be represented as a mass, spring system, as shown on Fig. \ref{fig:spring_mass}. 
For our specific case, we also introduce a tendon driven finger: a system of joints with antagonistic springs providing passive restoration to the resting position as shown in Fig. \ref{fig:finger_real_sim}. 

The ideal system dynamics of the tendon system can be written as:
\begin{equation}
    F_m - F_s - F_f = m_\text{eff}\ddot{x}
    \label{eq:force_balance}
\end{equation}
where $F_m = \tau / r$ is the motor force produced from motor torque $\tau$ and spool radius $r$; $F_s = k\Delta x$ is the spring restoring force with stiffness $k$ and displacement $\Delta x$; $F_f$ is the total friction force of the motor; $m_\text{eff} = J / r^2 + m$ is the sum of actuator and load inertia. 
For dexterous manipulators, the mass and acceleration of the system are significantly lower than the force and friction values, i.e. $m_\text{eff}\ddot{x} << F_m - F_f$.

\begin{figure}[tb]
    \centering
    \includegraphics[width=\linewidth]{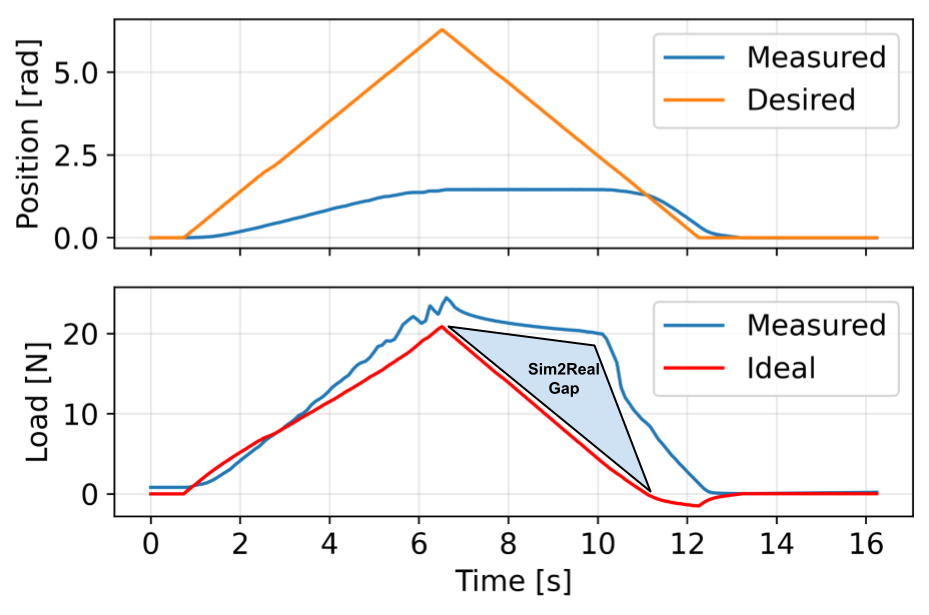}
    \vspace{-4mm}
    \caption{Motor motion that extends the spring and comes back to default position. Due to phenomena such as motor stick friction, the forces applied by the motor are non linear and result in high sim to real gap when ideal torque assumption is made in RL setting.}
    \label{fig:ramp}
    \vspace{-4mm}
\end{figure}

For typical consumer grade servo motors, such as widely used Dynamixels, motor torques are not directly available.
Actuation is typically generated through position control and tendon force is not known. 
Motor current can be directly controlled or measured, providing approximate torque sensing or control, however, this does not account for actuator friction or gearbox nonlinearities.
Motor friction, tendon slack, control delays and non-rigid body compliance of the system all contribute to force estimation error. 
% Friction is one of the biggest causes of error in this torque estimation as the gearbox friction is both direction and load-dependent. 
To demonstrate this error, we use our test bench shown in Fig. \ref{fig:spring_mass}, with an actuator and spring inline, and apply an open-loop position control ramp.  
% We see that that error between the desired and actual motor positions change depending on the applied load as can be observed in Fig. \ref{fig:ramp}. 
Using position data, we make an ideal force $\bar{F}$ source assumption of tendon force based upon the position control by scaling the position error between desired position $\theta^d$ and measured position $\theta$ by a gain $P$, which we empirically calculated to be 4.2 [N/rad] using: 

\begin{equation}
    \bar{F} = P(\theta^d - \theta),
    \label{eq:ideal}
\end{equation}

\noindent We see a significant 'gap' between this ideal and measured tendon force (Fig. \ref{fig:ramp} bottom).
This 'gap' or 'error' corresponds to sim2real gap when position control trajectories generated in simulation are transferred to the real world.
Whilst this behavior can be somewhat captured via simplified load-dependent models, such as in \cite{DuclusaudExtendedFrictionModels2025}, these models focus on capturing solely the friction characteristics of the motor.
We want to capture all causes of discrepancies from the ideal force source such as frictions, tendon slack, control delays, and non-rigid body compliance to minimize sim2real gap.
To capture this information, we propose to extend the data-driven approach of \cite{HwangboLearningAgileDynamic2019} to tendon driven system by leveraging history of measurements and its contextual information, enabling us to learn all motor-specific causes of discrepancy compared to an ideal force source.

% A simplified load-dependent model~\cite{DuclusaudExtendedFrictionModels2025} can explain this difference in behavior between the two motor modes, and is formulated as follows:
% \begin{equation}
%     F_f = K_v\,|\dot{s}| + K_c + K_\ell\,|F_m - F_e|
% \end{equation}
% where $K_v$ is the viscous friction coefficient, $K_c$ is the Coulomb friction coefficient, $K_\ell$ is the load-dependent friction coefficient, and $|\dot{s}|$ is the motor velocity. Given that $F_m$ and $F_e$ are the motor and external forces, it follows that the effective friction increases if the motor load increases, impacting the measured position error when the motor is in backdrive mode: load applied to the motor gearbox makes moving the gear teeth harder by increasing static friction. However, this approach does not capture all discrepancies between ideal and real model. Our approach learns the friction behavior along with any other cause of misalignment from data and maps the resulting effective tendon load rather than fitting a specific friction model. When the system is static, the tendon load is approximately:
% \begin{equation}
%     F \approx k\,q
% \end{equation}
% where $k$ is the finger torsional spring stiffness and $q$ is the joint displacement in radians.

\subsection{Data-Driven Tendon Force Modeling}
\label{sec:tendon_model}

Unlike in \cite{HwangboLearningAgileDynamic2019} where the actuators are highly dynamic and only required short window of history to estimate the motor forces, servo motors used in tendon-driven systems are much slower, requiring a longer contextual history window to capture the model discrepancies such as the error shown observed in Fig. \ref{fig:ramp}.
To address this, we propose multiple model architectures to address long history windows: a Multi Layer Perceptron with concatenated history vector, a Recurrent Neural Network and a Transformer architecture. 

All three models share the same type of inputs: a history of $H$ steps of observations $h_t$ at frequency $f$ [Hz]:

\begin{equation}
    o_t = [h_{t-H/f}, \,... \,h_{t-1/f},h_{t}]
    \qquad
    h_t = [\theta^d, \theta, \dot{\theta}]
    \label{obs}
\end{equation}

\noindent where $\theta^d$ is desired motor positions, $\theta$ is measured motor positions, $\dot{\theta}$ is measured motor velocities.
% The history corresponds to a timespan $T$ [s] at frequency $f$ [Hz]. 
This temporal context should be chosen to include enough information about the motor direction-dependent behavior, as shown in Fig. \ref{fig:ramp}.
In our test setup, we chose the $H$ and $f$ to be 30 and 20[Hz], respectively, for all three models, resulting in a history length of 1.5 [s].
The trained models provide a single scalar output corresponding to the estimated tendon load $\hat{F}$ in Newtons:

\begin{equation}
    \hat{F} = f_{model}(o_t)
    \label{est}
\end{equation}

\subsubsection{Multi Layer Perceptron}

The multi layer perceptron (MLP) takes the history window of $H$, flattened into a vector $\mathbf{x} \in \mathbb{R}^{3H}$, and maps it to the force estimate through a feed-forward network.

% with parameters $\boldsymbol{\theta}_\text{MLP}$:
% \begin{equation}
%     \hat{F} = f_{\boldsymbol{\theta}_\text{MLP}}(\mathbf{x}),
%     \label{eq:mlp}
% \end{equation}

The MLP consists of three hidden layers of 256, 128, 64 units respectively with ReLU activations followed by a linear output layer. The MLP is the lightest architecture and serves as a baseline.

\subsubsection{Recurrent Neural Networks}
% \label{sec:rnn}

The second model is a Recurrent Neural Network (RNN), which processes the input sequence step-by-step while maintaining a hidden state $\mathbf{h}_t \in \mathbb{R}^{64}$ that summarizes past information \cite{YuReviewRecurrentNeural2019}.

% \begin{equation}
%     \mathbf{h}_t = \tanh\!\left(\mathbf{W}_{hh}\mathbf{h}_{t-1} +
%     \mathbf{W}_{xh}\mathbf{x}_t + \mathbf{b}_h\right),
%     \label{eq:rnn}
% \end{equation}
% where $\mathbf{x}_t \in \mathbb{R}^3$ is the input at timestep $t$, $\mathbf{W}_{hh} \in \mathbb{R}^{64 \times 64}$ and $\mathbf{W}_{xh} \in \mathbb{R}^{64 \times 3}$ are the recurrent and input weight matrices, respectively, and $\mathbf{b}_h \in \mathbb{R}^{64}$ is the bias vector. 

At inference time, the hidden state is updated recurrently at each timestep and the current estimate $\hat{F}_t$ is obtained by passing $\mathbf{h}_t$ through a linear layer.
In theory, this allows the RNN to have indefinite history window.

\subsubsection{Transformer Encoder}
\label{sec:transformer}
The third model is a Transformer encoder~\cite{VaswaniAttentionAllYou2023}, which processes the entire input sequence $o_t \in \mathbb{R}^{H \times 3}$ in parallel through self-attention. 
The sequence is first projected to a 16-dimensional space via a linear layer, and sinusoidal positional encoding is added for temporal order information. 
The encoded sequence is then passed through 2 Transformer encoder layers, each with 4 attention heads. 
A causal mask is applied such that each timestep only considers past and current inputs, therefore the model is suitable for deployment on real hardware. 
The output at the final timestep is passed through a three-layer MLP to produce the force estimate $\hat{F}$.

\subsection{Data Collection}
\label{sec:data_collection}
To collect data for model training, we propose a data-collection test-bench which can generalize to different actuators and different robotic systems as shown in Fig. \ref{fig:spring_mass}.
The motor is connected via tendon to a mass-spring system which interfaces to a load cell allowing for recording of ground truth tendon force $F_\ell$ directly. 
To learn a motor model, we collect tendon force data when sending position commands $\theta^d$. 
We record motor position $\theta$ and velocity $\dot{\theta}$ from its encoder, and force from the load cell $F_\ell$ at a fixed rate of $80$\,Hz.

In terms of our system Eq. \ref{eq:force_balance}, the load cell force sensor measurement $F_\ell$ corresponds to tension provided by the spring or a finger of a dexterous hand, $F_s = F_\ell$.

\begin{equation}
    F_\ell = F_s = F_m - F_f - m_\text{eff}\ddot{x}
    \label{eq:load_cell}
\end{equation}

To capture training data that can enable the motor model to generalize over different robotic systems and also our robotic finger, we propose a mounting mechanism that allows for installation of a robotic finger used in the dexterous hand as shown in Fig \ref{fig:spring_mass}.
This allows us to collect data from the simplified spring-systems as well as our fingers, capturing some of the actuator behavior unique to control of tendon-driven system.
Data is collected with multiple spring and finger configurations to ensure that the model learns motor behaviors of various systems and becomes robot agnostic.
To teach the model contact-rich behavior, we blocked the finger at various positions during data collection replicating scenarios of the robot gripping items.
An example of such contact rich behavior is shown in Fig. \ref{fig:overview} where the finger holds an object.
The training data consists of random step responses ranging from [0.0, 2$\pi$], sinusoid with amplitude $\pi/2$ and frequency ranging within [$\pi$, 4$\pi$], ramps with slopes in [1, 2] range, and step stairs with 0.1 increment.
Such range of trajectories allows the model to experience wide range of motor dynamics.
% The training data motions consists of various steps, ramps and random-frequency sinusoids such the model can learn the system dynamics.
The range of actuator positions was selected to go from minimal forces applied by the motor to maximum force that the motor can apply when the finger is fully closed and undergoing self-contact.

\subsection{Simulation of Tendon Force-Driven Systems}
\label{sec:sim}

\begin{figure}[tb]
    \centering
    \includegraphics[width=\linewidth]{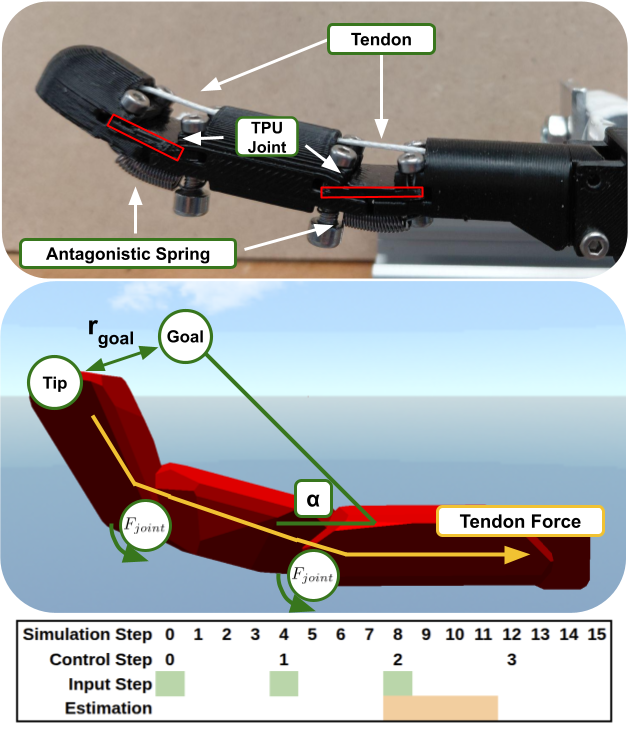}
    \vspace{-4mm}
    \caption{Finger test bench setup. The two coupled joints are made out of flexible TPU material. The finger is driven by a one tendon. Antagonistic springs are attached to the back of the finger to allow for release mechanism with tendons are not engaged. Example of simulated tendon driven finger. Spring forces are applied at joints. $\alpha$ is an example desired angle used for the RL controller. The force estimated by the model is calculated at control frequency and is applied to multiple consecutive simulation steps.}
    \label{fig:finger_real_sim}
    \vspace{-4mm}
\end{figure}

To simulate our test bench finger, we use a  rigid-body simulation capable of tendon force-driven systems as shown in Fig \ref{fig:finger_real_sim}.
At the joint level, we simulate the antagonistic force by applying a torque force generated by the attached spring elements with spring constant $k$: $F_{joint} = -k\Delta q$.
For actuation of the finger, we estimate and apply tendon force $\hat{F}$ using one of our models mentioned in (\ref{sec:tendon_model}).
To conserve computational resources and speed up the training process, we calculate the tendon forces at the control frequency, which is slower than the simulation frequency and apply it for multiple simulation time steps as shown in Fig. \ref{fig:finger_real_sim}.
Since there is no tendon spooling motor in simulation, we calculate the motor positions and velocities from tendon positions $l$ and velocities $\dot{l}$ via the motor spool radius $r$.
\begin{equation}
    \theta = l / r
    \qquad
    \dot{\theta} = \dot{l} / r
    \label{eq:tendon_to_force}
\end{equation}

\subsection{Reinforcement Learning}
\label{sec:RL}

The control problem is modeled as a Markov Decision Process (MDP) defined by the tuple $(\mathcal{S}, \mathcal{A}, P, \rho_0, r, \gamma)$, where the state space is assumed to be fully observable. 
At timestep $t$, the system provides features $o_t = [\theta \; \dot{\theta} \; a_t \; a_{t-1} \; \alpha]^T \in \mathbb{R}^{5}$, containing motor position $\theta$, motor velocity $\dot{\theta}$, current and previous actions $a_t$, $a_{t-1}$ and goal angle $\alpha$.
Goal angle corresponds to a desired end effector position of the tip as shown in Fig \ref{fig:finger_real_sim}.
The policy observes a temporal state history $s_t = (o_{t-k}, \ldots, o_{t-1}, o_t) \in \mathcal{S} \subseteq \mathbb{R}^{5k}$ with $k=30$ time-steps. 
The action space is $\mathcal{A} \subseteq \mathbb{R}$, containing motor position offset commands $\Delta\theta$ in radians.
The desired position is the sum of current position and the output of the controller:

\begin{equation}
    \theta^{d} = \theta + \Delta\theta
    \label{eq:rl_action}
\end{equation}

The desired position $\theta^{d}$ is then used to estimate tendon force $\hat{F}$ for simulation roll-outs.
We define $P(s_{t+1}|s_t, a_t)$ as the transition dynamics determined by the simulator, $\rho_0$ as the initial state distribution, $r: \mathcal{S} \times \mathcal{A} \to \mathbb{R}$ as a stationary reward function of states and actions, and $\gamma \in [0,1)$ as the discount factor.
The policy $\pi_\phi: \mathcal{S} \to \mathcal{P}(\mathcal{A})$ with parameters $\phi$ outputs a Gaussian distribution over actions.

To improve sim-to-real transfer, we use domain randomization~\cite{PengSimtoRealTransferRobotic2018}. During RL training, we randomize joint friction coefficients, link masses, spool radius and joint springs all within 30\% of the nominal values.

Let $\boldsymbol{\xi} \in \Xi \subseteq \mathbb{R}^4$ denote the environment parameters (joint friction coefficients, link masses, spool radius and joint springs), where $\Xi$ is the parameter space of randomized parameters.
The policy is trained by sampling $\boldsymbol{\xi} \sim p(\boldsymbol{\xi})$ from a distribution over $\Xi$, and the objective becomes:

\begin{equation}
    \max_{\phi} \mathbb{E}_{\boldsymbol{\xi} \sim p(\boldsymbol{\xi})} \left[ J(\pi_\theta; \boldsymbol{\xi}) \right]
    \label{eq:objective}
\end{equation}

\noindent where $J(\pi_\phi; \boldsymbol{\xi}) = \mathbb{E}_{\boldsymbol{\tau} \sim p(\boldsymbol{\tau} | \pi_{\phi}, \boldsymbol{\xi})} \left[ \sum_{t=0}^{H-1} \gamma^t r_t(\mathbf{s}_t, \mathbf{a}_t) \right]$ is the expected discounted return of policy $\pi_\phi$ with simulator parameters $\boldsymbol{\xi}$, discount factor $\gamma$ and $\boldsymbol{\tau} = (\mathbf{s}_0, \mathbf{a}_0, \mathbf{s}_1, \mathbf{a}_1, \ldots, \mathbf{s}_{H-1}, \mathbf{a}_{H-1})$ is a state-action trajectory. 
The policy learns to be less sensitive to specific values, and transfers more easily to the real system with parameters $\boldsymbol{\xi}_{\text{real}}$, which are unknown but assumed to lie within $\Xi$.

For the RL algorithm, we utilize PPO with modifications to the entropy coefficient, where we turn off all gradient flow to the gaussian standard deviation parameters \cite{SchulmanProximalPolicyOptimization2017, EngstromImplementationMattersDeep2020}. 
This results in reliable training where the exploration phase does not immediately collapse or increases significantly.

% \subsection{Controller Transfer to Real world}

The controller policy is trained using samples from simulation actuated by one of the three tendon force models specified in (\ref{sec:tendon_model}) and is then directly deployed on the real finger test bench in Fig. \ref{fig:spring_mass} to track the end effector position provided by the user via the $\alpha$ desired angle command shown in Fig. \ref{fig:finger_real_sim}.

The main rewards are: 

\begin{equation}
   r_{goal} = -||p_{goal} - p_{tip}||, \quad r_{smooth} = -||a_t - a_{t-1}||
    \label{eq:rew}
\end{equation}

\noindent  where $p_{goal}$ and $p_{tip}$ are goal and tip pose as shown in Fig. \ref{fig:finger_real_sim},
The controller network is a simple MLP with weights [128, 64, 32] and a RelU activation functions.

As a benchmark to compare our tendon force estimation against, we also train an RL policy using an ideal force source $\bar{F}$ described in Eq. \ref{eq:ideal}.

% provide an ideal tendon force source by calculating an empirical value of 4.2 [N/rad]. 
% And example of ideal tendon force is shown on Fig. \ref{fig:ramp}.

% \begin{equation}
%     \bar{F} = P(\theta^d - \theta),
%     \label{eq:ideal}
% \end{equation}

%% file: 03_experimental_setup.tex
\label{sec:experimental_setup}

\subsection{Test-bench \& Finger Setup}
The testbench system is a single degree-of-freedom (DoF) tendon-driven mechanism actuated by a Dynamixel Servo Motor XC330-T288-T servo with a spool radius of $r = 12.5$ mm.
A single tendon routed from the spool couples the proximal interphalangeal (PIP) and distal interphalangeal (DIP) joints, reproducing a coupled flexion motion similar to the human finger \cite{JungeADAPTTeleopRoboticHand2025}.
An antagonistic return spring provides passive extension, so that releasing tendon tension allows the finger to open without a second actuator.
Additionally, the revolute joints of the test bench finger are created via a 3D printed TPU element.
The TPU element provides additional spring element to the system.

For data collection and RL controller inference, we use a Raspberry Pi 5 running PyTorch.
The load cell is a Purecrea 5 kg Load Cell Weight Sensor read by a HX711 analog signal amplifier at 20 Hz and then extrapolated to 80 Hz to increase training data and allow for potential estimation at 80 Hz.
The motor signals measured position $\theta$ and measured velocity $\dot{\theta}$ are read at 80 Hz from the servo motor.
All data is recorded via a ROS2 system and rosbags.

Approximately a total of 36 [min] of data at 80 [Hz] was collected for this work. The split of data was approximately 50\%, 25\%, 25\% for finger test bench, weak spring and strong spring, respectively.
Inputs and outputs are independently normalized before training, and the normalization statistics are stored as buffers inside the exported model, such that inference does not require additional pre-processing.
All three models utilize similar number of parameters for fair comparison.
The tendon force estimation networks ranged from 0.1 to 0.2 [MB] in size.

\subsection{Simulation Environment}
To simulate the test setup, we use Vsim, a GPU-accelerated simulator that supports tendon force driven actuation ran on an RTX 5090 \cite{VsimNextgenPhysics}.
For the training algorithm we utilize "rl\_games", an open-source PPO implementation \cite{rl-games2021}.
We simulate the finger TPU joints as revolute joints and apply virtual motor force equal to the antagonistic spring force.

%% file: 04_results.tex
\label{sec:results}

In this section we first demonstrate the generalization of the models to different robot setups to demonstrate that we can train setup-agnostic actuator models that can be used for any RL training as long as the servo motor is the same.
We then perform a series of tests involving contact, to verify the models ability to simulate contact-rich scenarios.
To measure sim2real gap, we illustrate that using our model, sinusoidal trajectory that are performed using the forces from our model match closer to the real trajectories experienced on the test setup.
Finally, we compare the performance of a trained reinforcement learning policy using our model versus an estimated ideal tendon force source.

\subsection{Generalization of the Actuator Model}

\begin{figure*}[tbp]
    \centering
    \includegraphics[width=0.95\textwidth]{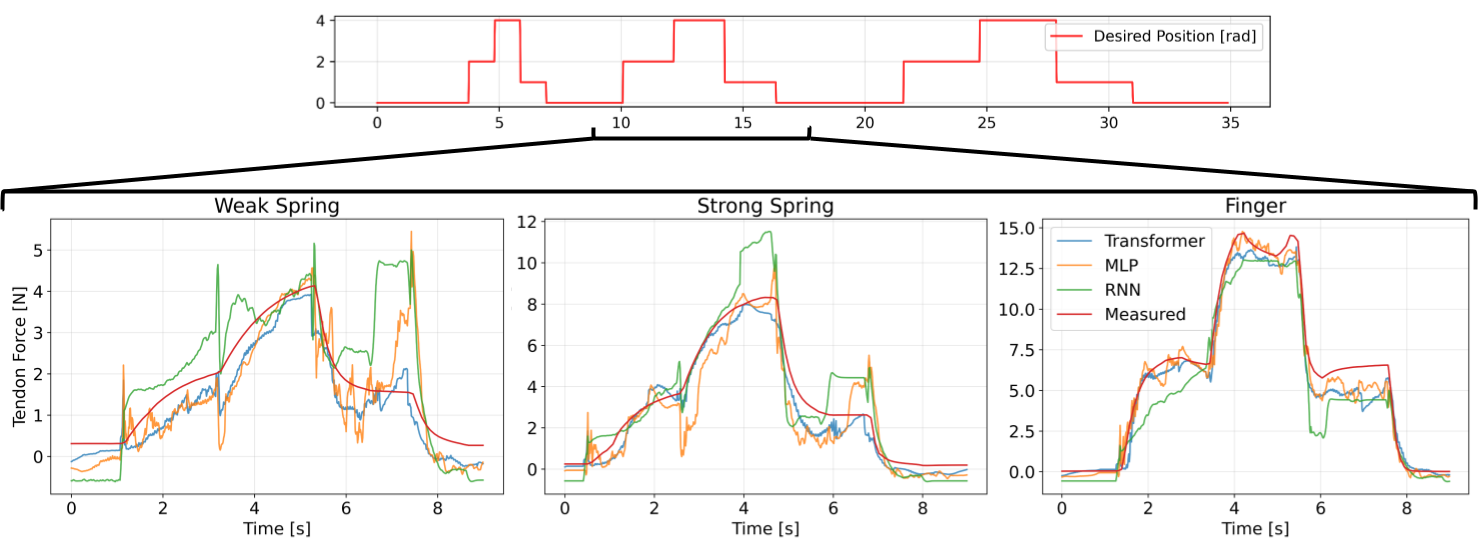}
    \vspace{-4mm}
    \caption{Prediction of tendon forces for a weak spring, strong spring and finger setup of a segment of the full testing trajectory corresponding to the 2.0 [s] steps as shown via the minimap. It can be seen that the transformer model manages to generalize to different measurements the best.}
    \label{fig:prediction_diff_springs}
    \vspace{-4mm}
\end{figure*}

\begin{figure}[tb]
    \centering
    \includegraphics[width=\linewidth]{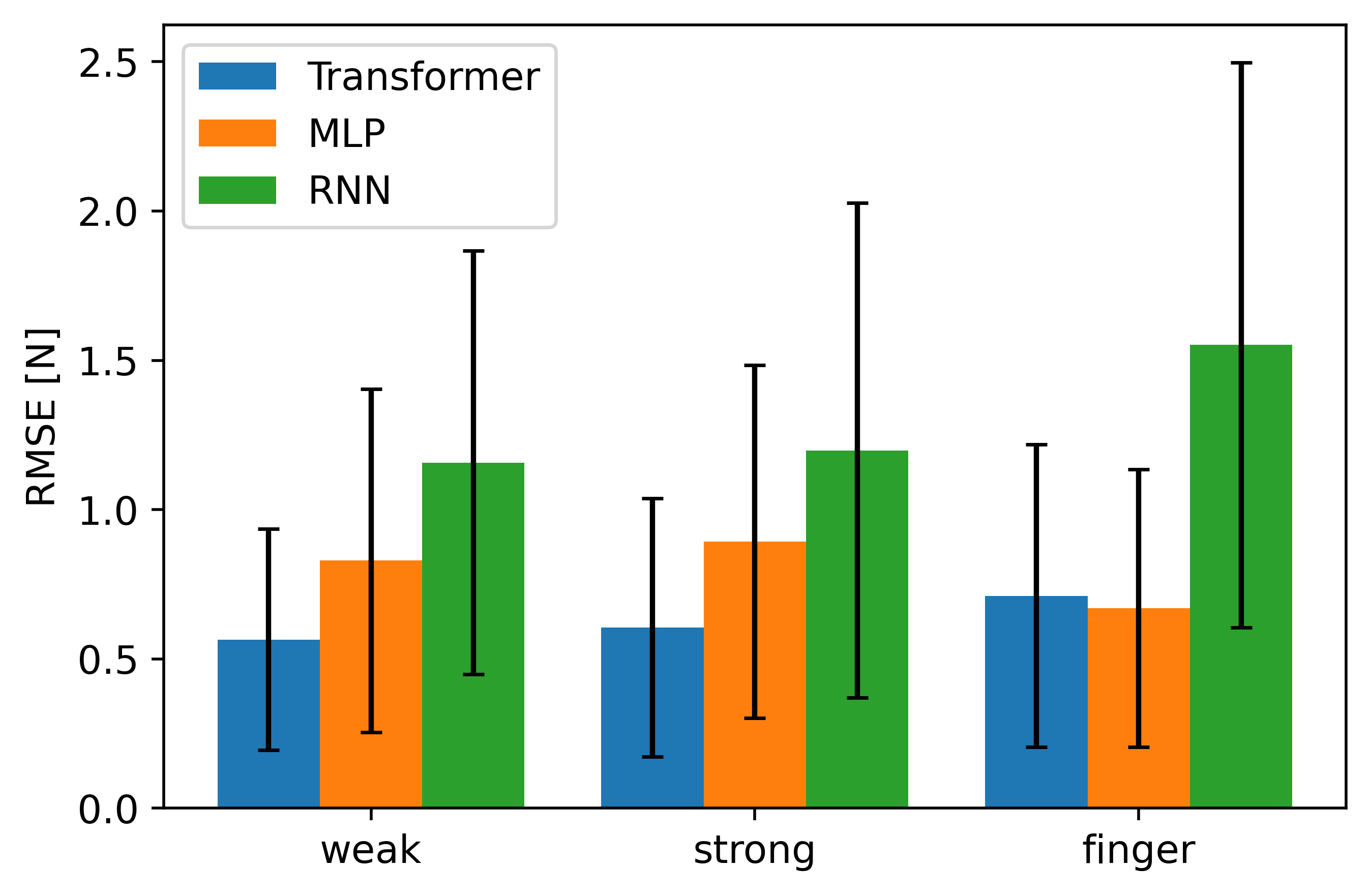}
    \vspace{-4mm}
    \caption{RMSE of tendon force estimation for various springs and models over the full trajectory of steps. The error bars represent standard deviation of absolute error.}
    \label{fig:rmse_bars}
    \vspace{-4mm}
\end{figure}

To evaluate the actuator model's ability to generalize to various robot models, we perform three identical step trajectories of desired motor positions for the three setups consisting of a weak spring, strong spring and the finger as seen in Fig. \ref{fig:spring_mass}. 
This represent using the same actuator model to predict tendon forces for different tendon driven robots. 
For each configuration we actuate the tendon using position control, with {0.0, 2.0, 4.0, 1.0} [rad] steps made, with the position then are held for {1, 2, 3} [s].
For each trajectory we estimate the tendon force using the three models we trained: MLP, RNN and Transformer. 
We test for different hold time periods to study how the models perform within their trained window of 1.5 [s] (history of 30 steps at 20 Hz inference), just outside of the trained window with 2.0 [s], and finally a much longer window of 3.0 [s] when all contextual information is lost, unless we use an RNN with potentially indefinite history window.
The predicted tendon force trajectories for the 2 second hold are given in Fig. \ref{fig:prediction_diff_springs} alongside the measured ground truth recorded by the load cell.
For the full trajectory with all hold lengths we compute the RMSE, with the errors given for the three configurations in  Fig. \ref{fig:rmse_bars}.

Across all three experiments the RNN diverges significantly from the measured forces, also predicting spikes in tendon force prediction when there is a step change in position.
The MPL and Transformer better track the true force, however, the MLP prediction incorporates some higher frequency, oscillatory predictions in force data, most likely due to over fitting to the data, leading to these small perturbations when we see a step change in position.
For all models we see the lowest error for the weak spring where the forces are lower, with the fingers results in relatively high error.
Comparing the RMSE and timeseries of the three different models we see that transformer and MLP both outperform the RNN which noticeably drifts.
While the MLP and Transformer models have similar performance on the finger setup, which had the largest chunk of training data present, we observe that the MLP model fails to generalize as well as the transformer model for the weak and the strong spring.
This fits our hypothesis about the MLP oscillating from small input perturbations.
This suggests that the transformer model has better generalizability compared to the other models.
The transformer model average RMSE over all systems is 0.61 [N] which corresponds to 2.9\% of the maximum force of 21 [N] that the servo motor is capable producing.

\subsection{Prediction of Contact Rich Trajectories}

To be useful for more dexterous interaction tasks, our model needs to be able to accurately estimate tendon forces during contact-rich scenarios. 
To test this, we performed a ramp up of desired motor position commands in cases with varying contact or self-collision.  This includes where we do not block the finger (the finger fully closes), blocking the finger half way with a semi-compliant material (a tape roll) and a scenario where we fully block the input using a human hand.

\begin{figure}[tb]
    \centering
    \includegraphics[width=\linewidth]{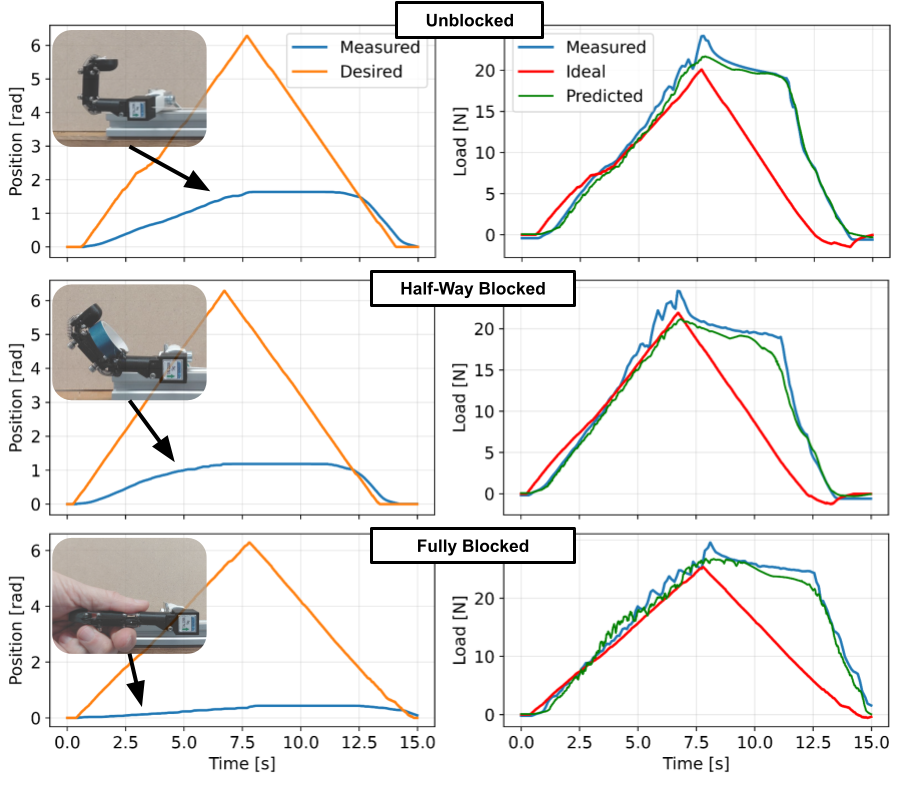}
    \vspace{-4mm}
    \caption{
    Transformer tendon model estimation of forces given unblocked, blocked half way via semi-rigid item and blocked at the start with a compliant human hand.
    The command is a ramp of desired motor position commands.
    The arrows indicate approximate locations where the finger was blocked, resulting in measured position to no longer change signficantly.
    We can see that in all situations, the transformer models predicts the correct tendon forces and motor nonlinearities.
    }
    \label{fig:pred_ramp}
    \vspace{-4mm}
\end{figure}

% RMSE Measured vs Predicted: 0.79 N
% RMSE Measured vs Predicted: 1.45 N
% RMSE Measured vs Predicted: 1.38 N

The results of the experiment are shown on Fig. \ref{fig:pred_ramp}. The measured versus predicted RMSE is 0.79, 1.45 and 1.38 [N] for the unblocked, partially blocked and fully blocked, respectively.
Qualitatively, we can see that in all three scenarios, the model predicts the measured tendon force well.
The ideal force source is accurate for the ramp up in position, as predicted, but fails to capture motor nonlinearities, and contact events.
This experiment demonstrates that our model is capable of capturing motor dynamics for contact-rich scenarios.

To further demonstrate the performance of the model during dynamic contact rich tasks, we perform a sinusoid desired motor position trajectory on the finger test bench and perturb the finger test by hand during its movement, simulating an object that is compliant or slipping during manipulation.

\begin{figure}[tb]
    \centering
    \includegraphics[width=\linewidth]{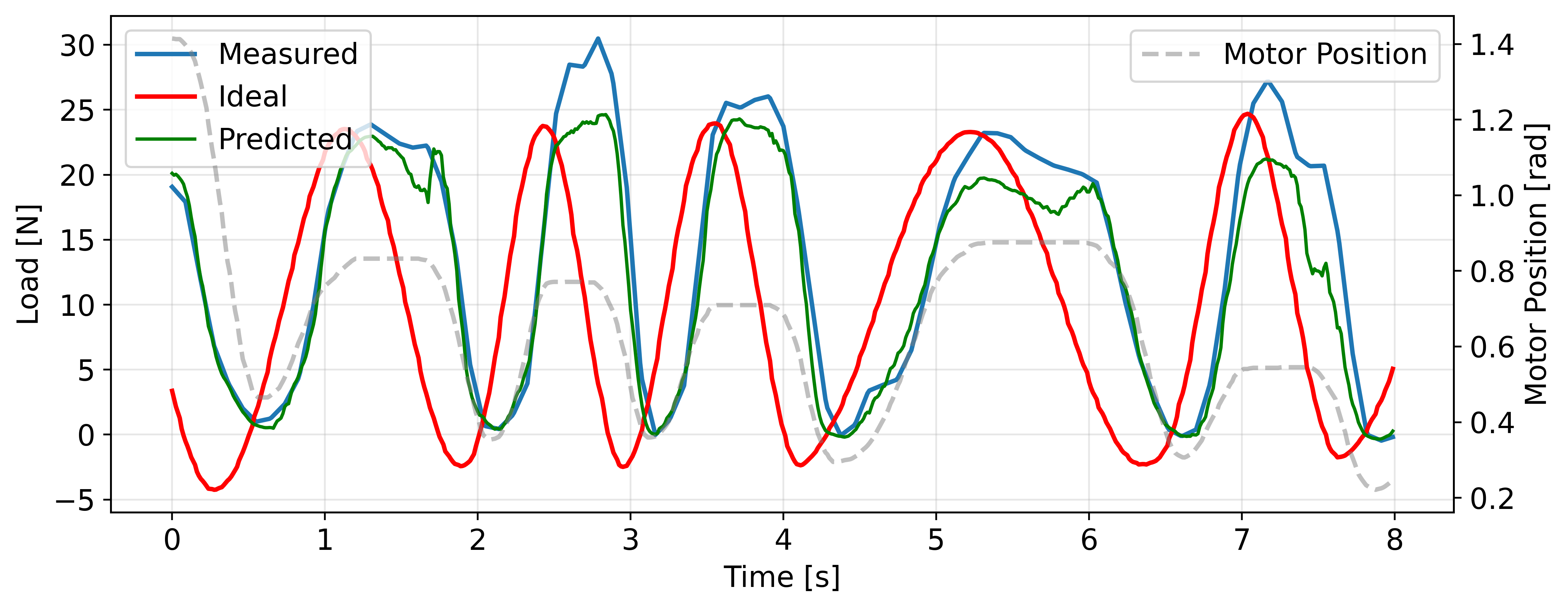}
    \vspace{-4mm}
    \caption{
    Transformer tendon force estimation during a sinusoidal motion of a the finger test bench.
    The finger was randomly perturbed by hand, simulating finger contacts and slipping.
    Real motor positions are visualized on the right axis, showing at which positions the finger was blocked.
    The transformer model captures the latency of the system, whereas the ideal force directly tracking the position error, resulting in temporal offset from the real tendon force.
    }
    \label{fig:pred_sine}
    \vspace{-4mm}
\end{figure}

% RMSE Measured vs Predicted: 2.08 N

The result is shown on Fig. \ref{fig:pred_sine}.
The RMSE of the measured versus prediction values is 2.08 [N], higher than when compared to the prediction error of the static or slow cases of Fig. \ref{fig:prediction_diff_springs} and \ref{fig:pred_ramp}.
We can see that the ideal source prediction is ahead of the real tendon force as it track the motor positional error, whereas as the transformer model is able to predict the latency of the output and nonlinearities in the behaviour.

\subsection{Sim2Real Gap Reduction with Tendon Force Model}

To quantify how the actuation model reduces the sim-to-real gap, we performed a sinusoidal position controller trajectory of the finger opening and closing on the real test setup and recorded the end effector position (tip) of the finger. 
The same position controlled trajectory was performed in simulation with tendon forces applied from our transformer model $F_{tendon}=\hat{F}$.
For comparison, we also modeled a tendon force model as a ideal force as in Eq. \ref{eq:ideal} with empirically calculated gain 4.2 [N/rad].

\begin{figure}[tb]
    \centering
    \includegraphics[width=\linewidth]{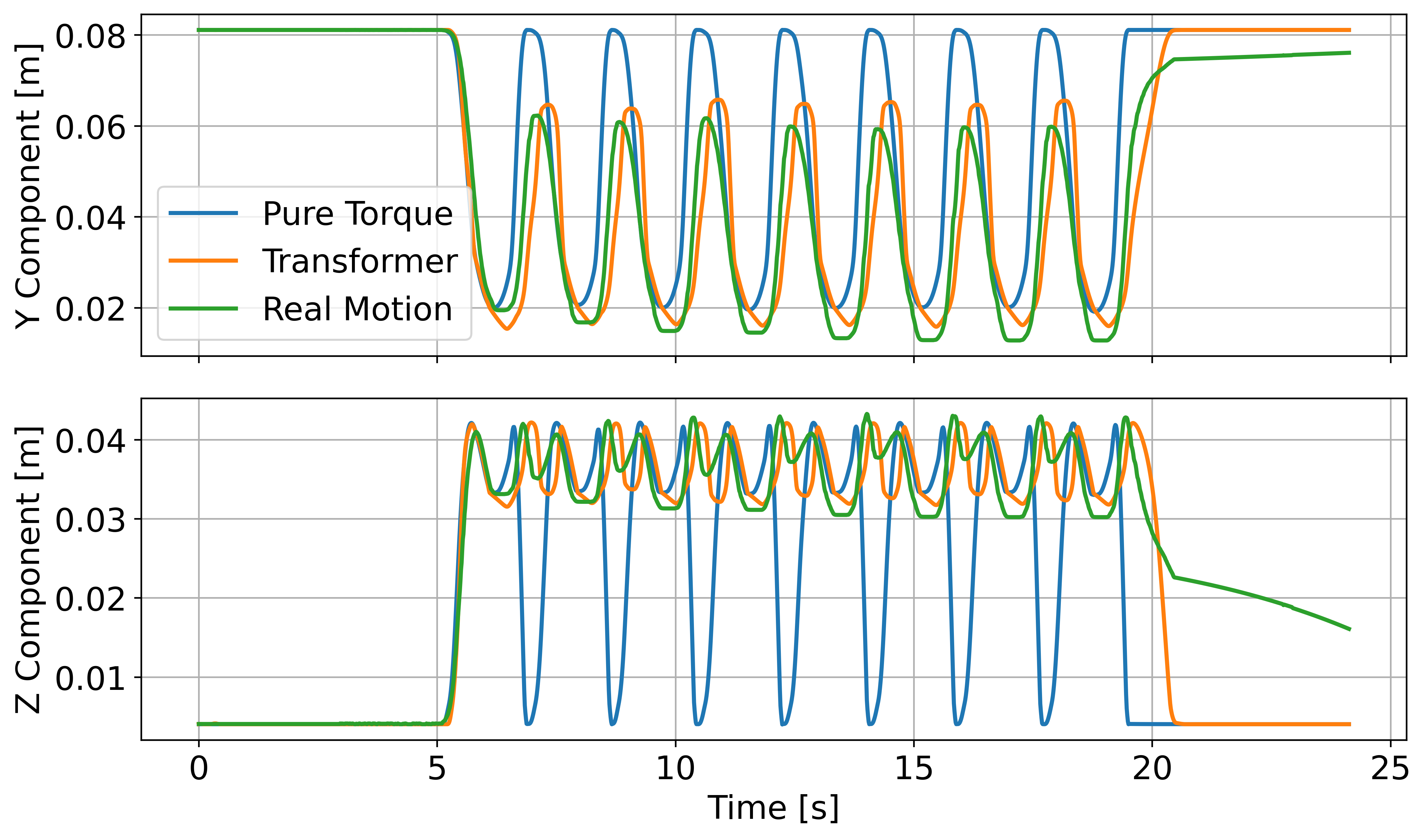}
    \vspace{-4mm}
    \caption{Finger tip position when simulating the desired position command through an ideal tendon force source, the transformer model, and the real motion recorded on the real finger.}
    \label{fig:sin_sim_vs_real}
    \vspace{-4mm}
\end{figure}

The resulting Fig. \ref{fig:sin_sim_vs_real} shows the tip position trajectory of the real motion, transformer model, and ideal tendon source.
The ideal tendon force source significantly overshoots and fully opens the finger, whereas the real motion and transformer motion are closer, with the transformer modulating the force to account for the friction and nonlinearities of the motor, so the finger does not fully open. 
The resulting RMSE for the transformer-based motion and ideal tendon force-based motion are 8.61 [mm] and 14.58 [mm], respectively. 
This results in a 41\% improvement in simulating a smooth sinusoidal trajectory.

\subsection{RL Policy Transfer to Real World}

To verify our models ability to produce reliable motion control RL policies that we can transfer to the real-world hardware, we train a controller to minimize the distance between the tip of the finger and a point along the finger arc, specified by an angle $\alpha$ as shown on Fig. \ref{fig:finger_real_sim}. 
We trained two controllers: one with the simulation that uses our transformer model to estimate tendon forces and one that utilizes an ideal tendon force source with Eq. \ref{eq:ideal} and gain set to 4.2 [N/rad]. 

% \begin{figure*}[htbp!]
\begin{figure}[tb]
    \centering
    \includegraphics[width=\linewidth]{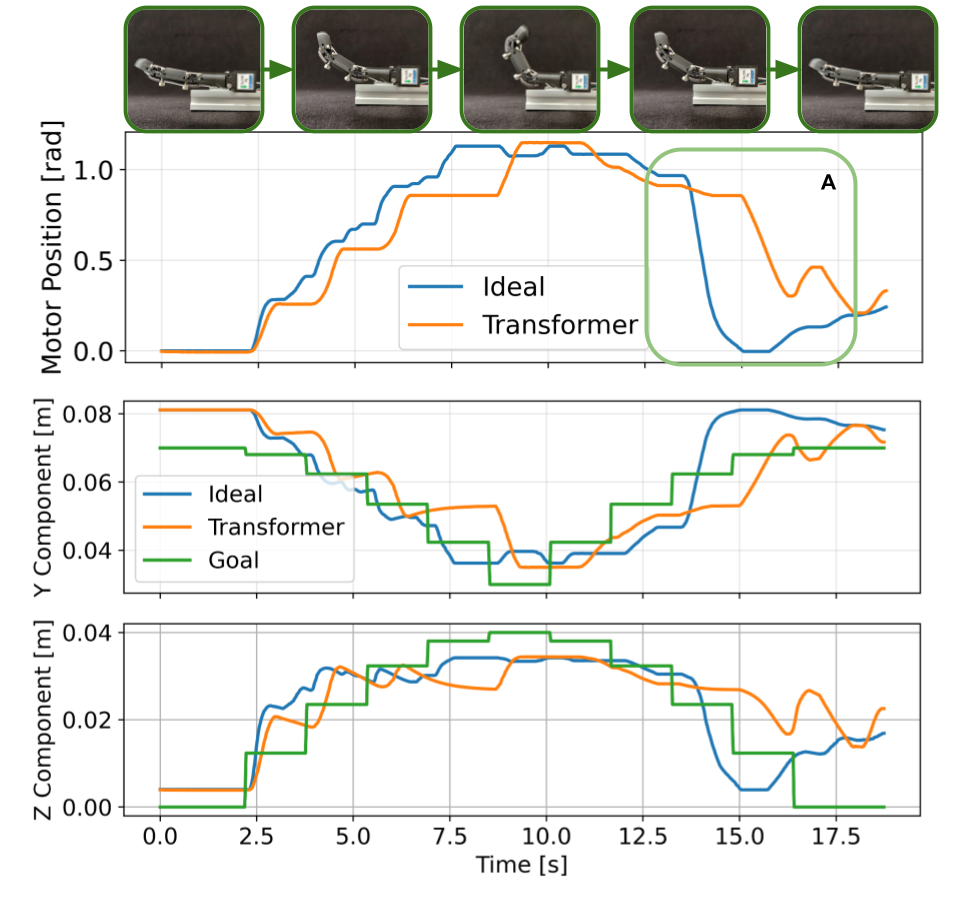}
    \vspace{-4mm}
    \caption{Plot of end effector tracking controller trained on ideal tendon force model and transformer-based model deployed on a real finger test bench. 
    Approximate finger motion is shown above. 
    Region A shows how the ideal tendon force model fails to track finger opening due to lack of knowledge of the motor nonlinearities, causing it to overshoot and fully open the finger.}
    \label{fig:rl_tracking}
    \vspace{-4mm}
% \end{figure*}
\end{figure}

The input target angle $\alpha$ was a stairs step varying from 0 to $\pi/2$ at $\pi/10$ increments, representing the finger closing with increasing $\alpha$.
The resulting trajectories of the end effector (the tip of the finger) are shown on Fig. \ref{fig:rl_tracking}. 
When curling the finger, both policies manage to track the desired end effector position well.
This matches what we have observed in Fig. \ref{fig:pred_ramp}, where the motion against the spring is well tracked by both ideal tendon force and the transformer model.
However, upon the release the finger ($\alpha$ is decreasing) in the region (A), we can see that the ideal tendon force trained policy overshoots, similarly to Fig. \ref{fig:sin_sim_vs_real} and fully opens the finger.
The transformer motion opens the finger slowly, minimizing the error between the tip and the goal pose.
The resulting RMSE for the transformer-based motion and pure torque-based motion are 12 [mm] and 24 [mm], respectively. 
Resulting in 50\% improvement in RL tracking performance when trained with our transformer model.

%% file: 05_discussion.tex
\label{sec:discussion}

Most tendon-driven dexterous hands use off-the-shelf servo motors that often lack any torque control and have high nonlinearities during control due to phenomena such as motor friction.
To leverage RL to develop robust, generalizable control polices we need to minimize the sim-to-real gap caused by these nonlinearities and enable deployment of force-aware controllers.
We provide a method to train a robot-agnostic, generalizing motor model that can predict tendon forces given a history of measured and desired positions and velocities.
We achieve this by implementing neural network models that are capable of extracting contextual information from a history of observations.
To train this model, we propose a new test bench which can directly collect data from the dexterous hand fingers.
Furthermore, by estimating the tendon forces, we can deploy this model in tendon-force driven GPU-accelerated simulators, enabling us to train RL policies with minimal sim2real gap when deploying on the real system.

We showed that the trained models are capable of capturing the motor dynamics and predicting reliable tendon forces.
We used the tendon estimation model to simulate one finger, a basic component of a tendon-driven dexterous hand, opening up the possibility of training RL controllers for hands built using compact servo motors.
Since the models are robot- and system-agnostic, one could train a library of these models for the most popular servo motors and provide a modular sim-to-real minimizing pipeline for any robot hand utilizing these motors.
This work provides a stepping stone for further research on reinforcement learning for tendon-driven dexterous hands.
For instance using this model to train a fully blind dexterous hand controller that relies on a history of proprioceptive measurements to perform manipulation tasks.
Furthermore, since we can predict forces applied by the tendons, the next step would be to use simulation data to train fingertip force predictions.